\documentclass[runningheads]{llncs}

\usepackage{eccv}
\RequirePackage[width=122mm,left=12mm,paperwidth=146mm,height=193mm,top=12mm,paperheight=217mm]{geometry}

\usepackage{eccvabbrv}

\usepackage{graphicx}
\usepackage{booktabs}

\usepackage[accsupp]{axessibility}  % Improves PDF readability for those with disabilities.

\usepackage{hyperref}

\usepackage{orcidlink}

\usepackage{bm}
\usepackage{bbm}
\usepackage{xcolor}
\usepackage{array,multirow}
\usepackage[linesnumbered,ruled,vlined]{algorithm2e}
\newcommand{\PAR}[1]{\noindent{\bf #1~}}
\definecolor{fgreen}{rgb}{0.13, 0.55, 0.13}
\definecolor{bred}{rgb}{0.55, 0.13, 0.13}

\DeclareMathOperator*{\argmax}{argmax}

\usepackage{rotating}
\newcommand{\lturn}[1]{\begin{turn}{90} #1 \end{turn}}

\begin{document}

% ---------------------------------------------------------------
% TODO REVIEW: Replace with your title
\title{Bayesian Self-Training for Semi-Supervised \\ 3D Segmentation} 

% TODO REVIEW: If the paper title is too long for the running head, you can set
% an abbreviated paper title here. If not, comment out.
\titlerunning{Bayesian Self-Training for Semi-Supervised 3D Segmentation}

% TODO FINAL: Replace with your author list. 
% Include the authors' OCRID for the camera-ready version, if at all possible.
\author{Ozan~Unal\inst{1,2}\orcidlink{0000-0002-1121-3883} \and
Christos~Sakaridis\inst{1} \and Luc~Van~Gool\inst{1,3,4}}

% TODO FINAL: Replace with an abbreviated list of authors.
\authorrunning{O.~Unal et al.}
% First names are abbreviated in the running head.
% If there are more than two authors, 'et al.' is used.

% TODO FINAL: Replace with your institution list.
\institute{$^1$ETH Zurich, $^2$Huawei Technologies, $^3$KU Leuven, $^4$INSAIT \\
\email{\{ozan.unal, csakarid, vangool\}@vision.ee.ethz.ch}}

\maketitle

\begin{abstract}
    3D segmentation is a core problem in computer vision and, similarly to many other dense prediction tasks, it requires large amounts of annotated data for adequate training. However, densely labeling 3D point clouds to employ fully-supervised training remains too labor intensive and expensive. Semi-supervised training provides a more practical alternative, where only a small set of labeled data is given, accompanied by a larger unlabeled set. This area thus studies the effective use of unlabeled data to reduce the performance gap that arises due to the lack of annotations. In this work, inspired by Bayesian deep learning, we first propose a Bayesian self-training framework for semi-supervised 3D semantic segmentation. Employing stochastic inference, we generate an initial set of pseudo-labels and then filter these based on estimated point-wise uncertainty. By constructing a heuristic $n$-partite matching algorithm, we extend the method to semi-supervised 3D instance segmentation, and finally, with the same building blocks, to dense 3D visual grounding. We demonstrate state-of-the-art results for our semi-supervised method on SemanticKITTI and ScribbleKITTI for 3D semantic segmentation and on ScanNet and S3DIS for 3D instance segmentation. We further achieve substantial improvements in dense 3D visual grounding over supervised-only baselines on ScanRefer. Our project page is available at \url{ouenal.github.io/bst/}.
    
\end{abstract}

\section{Introduction}

Dense 3D semantic perception at the level of points is pivotal for several applications that are based on fine-grained processing and reasoning in 3D, e.g.\ autonomous cars, robotics, and augmented reality. Such perception requires a segmentation of the unstructured input 3D point cloud into different semantic classes, i.e.\ 3D semantic segmentation~\cite{zhang2020deep}, or distinct instances of these classes, i.e., 3D instance segmentation~\cite{lai2023mask}. The input point cloud is often coupled with human utterances that refer to individual objects in the former in cases where a visual AI system interacts with human users in a common 3D environment, necessitating a 3D instance-level segmentation of these objects based on the aforementioned referrals, a task which is known as dense 3D visual grounding or referral-based 3D instance segmentation~\cite{yuan2021instancerefer,unal2023three}. In the majority of works that address these diverse 3D segmentation tasks, it is assumed that the training data come with full 3D semantic and/or verbal annotations, which creates the pressing need for large-scale labeled 3D datasets~\cite{dai2017scannet,armeni20163d,chen2020scanrefer,achlioptas2020referit3d} so that large end-to-end models can be adequately fitted on these data.

\begin{figure}[t]
    \centering
    \includegraphics[width=\textwidth]{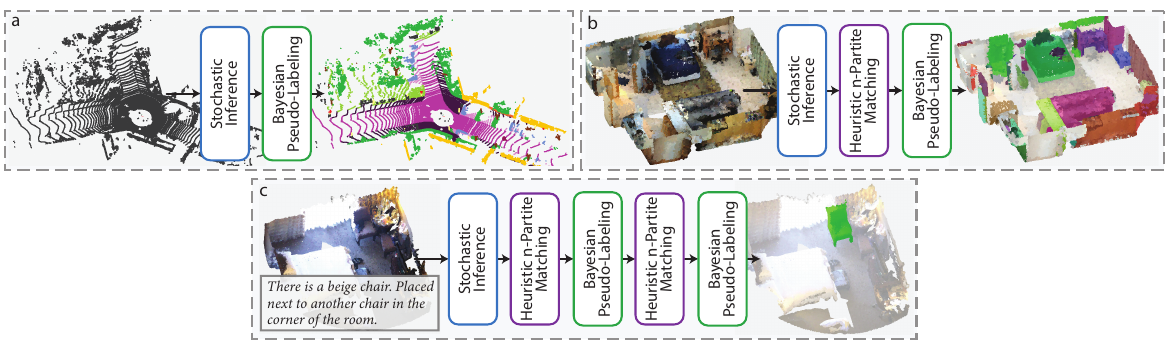}
    \caption{Illustration of our Bayesian pseudo-labeling pipeline for semi-supervised a) 3D semantic segmentation; b) 3D instance segmentation; and c) dense 3D visual grounding. With only slight adjustments, using the same building blocks, our method can be adapted to each of these tasks to achieve SOTA results.}
    \label{fig:teaser}
\end{figure}

\sloppy{However, the creation of such fine-level 3D annotations and verbal referrals for 3D scans requires human labor to be fairly reliable. Such labor is time-consuming and expensive, especially due to the high dimensionality of the scanned point clouds, which makes it hard for human annotators to navigate through them, visually inspect them, and verbally describe them. These practical difficulties render learning frameworks for 3D segmentation that operate with partial labels essential, semi-supervised learning being a primary case.
Answering this need, several recent works address semi-supervised 3D semantic segmentation~\cite{kong2023lasermix,jiang2021guided,cheng2021sspc}, most of which focus on effective pseudo-labeling techniques for self-training on the larger, unlabeled portion of the training data.
However, the generation of pseudo-labels in these works typically focuses on some grouping mechanism over 3D points, such as superpoints, to refine and propagate pseudo-labels and otherwise relies on na\"ive core techniques such as confidence thresholding~\cite{zou2018unsupervised} for generating the \emph{initial} point-level pseudo-labels that are then post-processed by their proposed modules.
Adding to this, there is a relative scarcity in works on semi-supervised 3D \emph{instance} segmentation~\cite{chu2022twist,liu2022weakly}, which we attribute to the relative difficulty in pseudo-labeling instances compared to creating plain semantic pseudo-labels. Last but not least, to our knowledge, no work has considered the setting of semi-supervised dense 3D visual grounding.}

We jointly address these shortcomings of previous works by proposing a common self-training framework for semi-supervised 3D segmentation which is based on Bayesian deep learning and applies to all three aforementioned segmentation tasks. At the core of our self-training framework lies a variational Bayesian approximation of the uncertainty of the respective network's predictions via dropout-based Monte Carlo integration~\cite{gal2016dropout} over multiple augmented versions of the original 3D input. This approximation delivers a more informed uncertainty estimate than simple confidence thresholding techniques~\cite{zou2018unsupervised}, which is crucial for eliminating confirmation bias from the pseudo-labeling process that we adopt for self-training. In particular, we utilize both the estimated Shannon entropy from the sampled predictions and the predicted label from each sample to filter out uncertain or inconsistent points from the generated set of pseudo-labels.

While this formulation is sufficient for the task of 3D semantic segmentation, it does not work out of the box for instance-level segmentation, as Monte Carlo integration cannot be applied directly to instance predictions. To account for this issue, we propose a {novel and} efficient, heuristic matching strategy across instances coming from different stochastic dropout forward passes based on linear sum assignment. This strategy enables the application of the same filtering steps as described above for semantic pseudo-labels to \emph{instance} pseudo-labels and thus extends the scope of our approach to 3D instance segmentation. Going one step further, we observe that the same matching strategy that we devise for instance segmentation applies to the selection stage of dense 3D grounding, in which the correct instance from the set of segmented candidate instances needs to be selected based on the input verbal description. Thus, for dense 3D grounding, we apply our heuristic linear sum assignment twice, first for the purely 3D-based candidate instance proposal generation and afterward for the verbo-visual instance selection. Our complete self-training pipeline for each of the three 3D segmentation tasks to which our method applies is summarized in Fig.~\ref{fig:teaser}.

We test our method on leading benchmarks across a wide range of domains and datasets. Specifically, our Bayesian self-training framework achieves state-of-the-art results for semi-supervised 3D semantic segmentation on SemanticKITTI~\cite{behley2019semantickitti} and ScribbleKITTI~\cite{unal2022scribble} and semi-supervised 3D instance segmentation on ScanNet~\cite{dai2017scannet} and S3DIS~\cite{armeni20163d}. Furthermore, we are the first to explore a semi-supervised setting on ScanRefer~\cite{chen2020scanrefer} for 3D visual grounding and obtain consistent significant improvements over supervised-only baselines.

\noindent In summary, our contributions are as follows:
\begin{enumerate}
    \item {Following established uncertainty estimation techniques from active learning,} we construct a Bayesian self-training framework for semi-supervised 3D semantic segmentation by generating reliable pseudo-labels through stochastic inference and entropy-based uncertainty filtering.
    \item We introduce a novel heuristic matching algorithm to extend the proposed framework to 3D instance segmentation.
    \item We further extend the method to dense 3D visual grounding utilizing the same building blocks as in the above tasks and are the first to explore this task in a semi-supervised setting.
    \item We obtain state-of-the-art results for semi-supervised 3D segmentation on SemanticKITTI~\cite{behley2019semantickitti}, ScribbleKITTI~\cite{unal2022scribble}, ScanNet~\cite{dai2017scannet} and S3DIS~\cite{armeni20163d}.
\end{enumerate}

% \clearpage
% \hphantom{o}
% \newpage

\section{Related Work}
\label{sec:related}

\PAR{Semi-supervised 3D semantic segmentation} has recently grown in popularity in the area of 3D understanding along with weakly supervised 3D semantic segmentation~\cite{unal2022scribble,wei2020multipath,xu2020weakly,zhang2021weakly}, due to the difficulty in large-scale 3D labeling that is necessary for the supervised setup~\cite{zhang2020deep}. {In this work, we tackle the task of semi-supervised segmentation where a large portion of data is completely unlabeled. As opposed to methods developed for weakly-supervised settings, a semi-supervised method cannot leverage the availability of partial true label information within each data sample, e.g. via the ArcPoint loss~\cite{lee2023gaia} or a PLS descriptor~\cite{unal2022scribble}. A common technique for leveraging unlabeled data under this restrictive scheme is explored under pseudo-labeling.} Semantic pseudo-labels are utilized by Jiang~\etal~\cite{jiang2021guided} to guide a contrastive loss and in SSPC~\cite{cheng2021sspc} to propagate labels on a superpoint graph. Superpoints are also leveraged by Deng~\etal~\cite{deng2022superpoint} to constrain pseudo-label propagation. In the 2D counterpart task, Chen~\etal~\cite{chen2021semi} enforces consistency of semantic pseudo-labels across perturbed versions of the 2D segmentation network. By contrast, we perturb both the network weights and the 3D inputs across different forward passes, and we consider both the entropy and the consistency among predictions in our pseudo-labels. Another related line of 2D semi-supervised works generates pseudo-labels based on direct thresholding of max-softmax scores to reduce confirmation bias~\cite{zou2018unsupervised,he2021redistributing}, while we threshold the \emph{entropy} of the distribution of softmax scores. Active learning is combined with a 3D semi-supervised approach in DiAL~\cite{unal2023discwise} to streamline practical 3D segmentation annotation pipelines. LaserMix~\cite{kong2023lasermix} introduces variability through cylindrical and range-view partitioning and mixing of 3D LiDAR scans, encouraging consistency across different mixes. % Language-based pre-training is employed in~\cite{chen2023clip2scene}, which pre-trains a 3D segmentation network via semantic and spatial-temporal consistency regularization.

\PAR{Semi-supervised 3D instance segmentation} is still in its infancy compared to its instance-agnostic semantic segmentation counterpart reviewed above. In particular, most object-level semi-supervised 3D understanding works focus on the coarser task of 3D object detection~\cite{tang2019transferable,wang20213dioumatch,zhao2020sess,gao2023dqs3d,wang2021semi,qin2020weakly,meng2021towards,ren20213d}, which outputs 3D bounding boxes that are not fine enough for precision-critical downstream tasks such as robotic grasping. For 3D instance/object segmentation, approaches that depart from the standard fully supervised setup have mostly considered the cases of unsupervised pre-training~\cite{xie2020pointcontrast}, weak supervision via one-point-one-instance~\cite{hou2021exploring,tao2022seggroup}, or unsupervised generic 3D object segmentation~\cite{lei2023efem}. To our knowledge, the only two semi-supervised works that specifically consider 3D instance segmentation are TWIST~\cite{chu2022twist} and WS3D~\cite{liu2022weakly}. The former work proposes a pseudo-label generation strategy tailored for the instance segmentation task, for which the technique of confidence thresholding that is ubiquitously used to the same end in semantic segmentation is not applicable. It thus introduces pseudo offset vectors, besides semantic proposal-level pseudo-labels, for performing self-training. The latter work proposes a region-level contrastive loss, which performs unsupervised instance discrimination, and leverages boundary information via a region-level unsupervised energy-based loss. On the contrary, we propose a unified semi-supervised Bayesian self-training framework that spans multiple 3D segmentation tasks including 3D instance segmentation, and handles the additional matching required between instances generated in different forward passes through the network via an efficient heuristic optimization step.

\PAR{3D visual grounding} is a central 3D task in the area of vision and language~\cite{prabhudesai2020embodied,feng2021free,luo20223d,kong2014coreference,zhao20213dvgt,cai20223djcg}. Relevant models accept a 3D point cloud as visual input and perform grounding of the accompanying verbal description to the referred object. While most previous works~\cite{chen2022ham,huang2022multiview,roh2022languagerefer,luo20223d,jain2022bottom} focus on coarse 3D bounding box outputs, the task of \emph{dense} 3D visual grounding or referral-based 3D instance segmentation has been explored far less in comparison, in which a fine-grained point-level mask is predicted~\cite{huang2021tgnn,yuan2021instancerefer,unal2023three}. Regardless, all aforementioned methods invariably assume a \emph{supervised} setting, where full ground-truth instance annotations are available for all samples at training time. By contrast, the formulation of our method makes it fit for also handling the more practical \emph{semi-supervised} setting, in which only a part of the training data is labeled.

\PAR{Uncertainty estimation} is crucial in learning setups that lack full supervision as ours, given that such estimates can provide information about the reliability of the predictions of the model on unlabeled samples~\cite{houlsby2011bayesian}. In the more recent deep learning era, Bayesian neural networks form the core for uncertainty estimation by modeling network weights as stochastic distributions, where test-time-dropout~\cite{srivastava2014dropout} is utilized as a variational Bayesian approximation~\cite{gal2016dropout} and Monte Carlo integration is computed via multiple stochastic forward passes to approximate prediction uncertainty~\cite{gal2017active}. This Bayesian deep learning framework has been applied successfully to supervised 2D dense prediction tasks~\cite{kendall2017bayesian, kendall2018multitask}. We focus on the aggregate uncertainty in the predictions of a general 3D segmentation model and build on the Bayesian deep learning approach of Gal~\etal~\cite{gal2017active} which examines the approximate predictive entropy as an acquisition function for active learning. However, contrary to Gal~\etal~\cite{gal2017active}, in our self-training setup, we use \emph{low} rather than high entropy as a positive indication for considering a point as a source of supervision. To the best of our knowledge, our work is the first to introduce such a Bayesian self-training framework for general 3D point cloud segmentation, {specifically including 3D instance segmentation and referral-based 3D instance segmentation}.

Related uncertainty-based approaches to ours have been proposed for the {semantic} segmentation of 3D grid volume data in the field of medical image computing~\cite{yu2019uncertainty,xia2020semi, dai2023semi}. In particular, Yu~\etal~\cite{yu2019uncertainty} also use Monte Carlo dropout to estimate predictive entropy but only applies dropout to a single network input, whereas we vary both network weights and input views. Moreover, that work leverages the estimated entropy merely as the weight in a consistency loss between a teacher and a student network, while we filter pseudo-labels based on our entropy estimates. Besides, Xia~\etal~\cite{xia2020semi} perform semi-supervised semantic segmentation with multi-view co-training and applies Bayesian deep learning across views for uncertainty-based weighting of pseudo-labels. We only train one network instead of multiple networks employed in Xia~\etal~\cite{xia2020semi} and we derive the uncertainty estimates by feeding all input views to that single network, which is more parameter-efficient. Both of these approaches only address 3D semantic segmentation, {and cannot be utilized in instance-level dense prediction tasks.}
%while we tackle a broader range of tasks.
% , including 3D semantic segmentation, 3D instance segmentation and referral-based 3D instance segmentation.

\section{Bayesian Self-Training}

\begin{figure}[t]
    \centering
    \includegraphics[width=\textwidth]{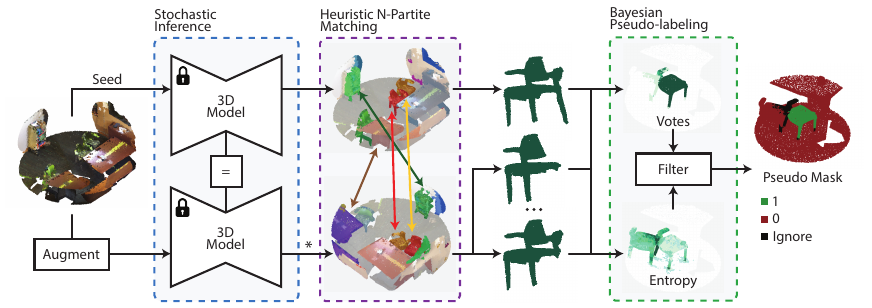}
    \caption{Illustration of Bayesian pseudo-labeling for semi-supervised instance segmentation. We initialize a seed prediction via a forward pass using a non-augmented input. Then our heuristic $n$-partite matching algorithm is employed to pair each seed instance with the best matching predicted instance from each of the $K$ stochastic forward passes. For each aligned object, we compute the aggregated label through unanimous voting and filter it based on uncertainty to obtain the final pseudo mask.}
    \label{fig:inst_pipeline}
\vspace{-10px} \end{figure}

In Sec.~\ref{sec:semi_sem} we start by formulating a self-training approach for semi-supervised 3D semantic segmentation {based on established uncertainty estimation techniques from the field of active learning~\cite{gal2016dropout,unal2023discwise}}. We then extend our formulation to include instance labels in Sec.~\ref{sec:semi_inst} and finally build up to also take natural language prompts as input, i.e., towards semi-supervised dense 3D visual grounding in Sec.~\ref{sec:semi_vg}.

\subsection{Semi-Supervised 3D Semantic Segmentation}
\label{sec:semi_sem}

In a semi-supervised setting, a dataset $D$ consists of a labeled set $S$ and an unlabeled set $U$. The baseline approach when working with such data is to treat the problem similarly to any fully supervised task and employ a semantic loss $\mathcal{L}_{sem}$ on the available labeled scenes $S$, typically stated as cross-entropy. Formally, the objective function over the model parameters $\omega$ is formulated as:
\begin{equation} \label{eq:partial_loss}
    \min_{\omega} \sum_{i=1}^{|S|} \mathcal{L}_{sem}(\bm{\hat y}, \bm{\mathrm{y}}).
\end{equation}

Starting from this baseline formulation, a semi-supervised training strategy aims to leverage the unlabeled scenes $U$ to further improve model performance. One such common technique is self-training, where the model is iteratively trained on its own predictions, i.e.\ on generated per-point one-hot pseudo-label vectors $\hat{\bm{\mathrm{y}}} = [\hat{\mathrm{y}}^{(1)}, \dots, \hat{\mathrm{y}}^{(C)}]$, with $C$ denoting the number of classes. However, the task of generating pseudo-labels is not trivial, as the model may suffer from \textit{confirmation bias} that occurs when repeatedly overfitting to incorrect labels. A simple but effective solution is explored under the umbrella of \textit{thresholding}~\cite{cascante2021curriculum,xie2020selftraining, zoph2020rethinking,zou2018unsupervised,he2021redistributing}. In essence, instead of training on all model predictions, a pseudo-label set $L$ is formed by predictions that exceed a confidence of a determined threshold $\tau$ to ensure a higher level of accuracy.
Formally, the objective function of threshold-based self-training for semantic segmentation can be formulated as:
\begin{equation} \label{eq:st}
\begin{split}
        \min_{\omega, \hat{\bm{\mathrm{y}}}}
            \sum_{d=1}^{|D|}
                \left( \vphantom{\sum_{r=1}^R}
                    \mathcal{L}_{sem}(\bm{\hat y}, \bm{\mathrm{y}}) - (\log \hat{\bm{y}}_{d} - \tau) \cdot \hat{\bm{\mathrm{y}}}_{d}
                \right) \\
\end{split}
\end{equation}
with $\hat{\bm{y}}_{d}$ denoting the softmax model predictions. Eq.~\ref{eq:st} is optimized in two steps that are repeated until convergence:
\begin{enumerate}
    \item \textbf{Training:} First, $\hat{\bm{\mathrm{y}}}$ is fixed, and the objective function is optimized with respect to $\omega$.
    \item \textbf{Pseudo-labeling:}  Then, $\omega$ is fixed and the objective function is optimized with respect to $\hat{\bm{\mathrm{y}}}$. Consequently, $L$ is updated given $\hat{\bm{\mathrm{y}}}$.
\end{enumerate}
The process is initialized by setting the latent variable $\hat{\bm{\mathrm{y}}} = \mathbf{0}$ for all points, i.e.\ by reverting to the na\"ive solution of Eq.~\ref{eq:partial_loss} and only training on the labeled set $S$. The two steps can be repeated to take advantage of the improved representation capability of the model thanks to pseudo-labeling.

While na\"ive thresholding relies on the softmax scores of the model to determine confidence, these scores may be unreliable and often skewed when facing long-tailed distributions, resulting in overconfident predictions~\cite{guo2017calibration}. Inspired by uncertainty estimation techniques commonly employed in active learning (AL) pipelines~\cite{houlsby2011bayesian,unal2023discwise}, we introduce a Bayesian self-training strategy that leverages an uncertainty-aware filtering technique for generating more reliable pseudo-labels. % Unlike in AL, we use the uncertainty metrics in the opposite manner, where instead of employing an oracle to label extracted uncertain points, we use the model to label confident points.

We first start by decomposing the goal of 3D semantic segmentation, which is to discover the dependency of the point-wise distribution over the labels $y \in Y$ on an input sample $x \in X$ via the model weights $\omega$. Formally, the task can be reduced to per-point classification, where $y$ can be of class $c \in \{1,...,C\}$. The conditional probability is therefore defined by:
\begin{equation}
    p(y=c|x,D)  \hspace{-.1em} = \hspace{-.3em} \int \hspace{-.3em} p(y=c|a(x),\omega) p(\omega|D) p(a(x)|x) d\omega dx
\end{equation}
for the dataset $D = (X,Y)$, under randomly applied affine transformations modeled through $a(\cdot)$. %

We set $p(a(x)|x)$ as a constant, restate the intractable $p(\omega|D)$ via the variational approximation $q(\omega)$ and minimize KL$(q(\omega)|p(\omega|X,Y))$. This objective can be approximated through Monte Carlo integration by employing variational inference via Bayesian networks, i.e.\ by applying stochastic forward passes with weights $\omega_k$:
\begin{equation}
\begin{split}
    p(y=c|x,D) & \approx \int p(y=c|a(x),\omega) q(\omega) d\omega dx \\
     & \approx \frac{1}{K}\sum_{k=1}^Kp(y=c|a(x),\omega_k).
    \label{eq:mc_integration}
\end{split}
\end{equation}
Compared to the common decomposition of BALD~\cite{houlsby2011bayesian}, the inclusion of the affine transformations via $a(\cdot)$ allows a higher deviation amongst stochastic forward passes which in return allows better estimation of predictive uncertainty~\cite{wang2019aleatoric}. We capture the uncertainty via the Shannon entropy~\cite{shannon1948mathematical}:
\begin{equation}
    \mathbb{H}[y|x,D] = - \sum_{c=1}^C p(y=c|x,D) \log p(y=c|x,D)
\end{equation}
and the accumulated predicted label $\hat{\bm{\textrm{y}}}^*$ is given by the unanimous voting principle:
% \begin{equation}
%     \hat{y}^{*(c)} = 
%     \begin{cases}
%     1\textrm{, if } K = \max \sum_{k} \Lim{h\rightarrow \infty} \frac{\exp(p(y|a(x),\omega_k))^h}{\sum_c \exp(p(y|a(x),\omega_k))^h} \\
%     0\textrm{, otherwise.}
%     \end{cases}
%     \label{eq:mc_label}
% \end{equation}
\begin{equation}
    \hat{\textrm{y}}^{*(c)} = 
    \begin{cases}
    1\textrm{, if } K = \sum_{k} \mathbbm{1}(\argmax (p(y|a(x),\omega_k)) = c)\\
    0\textrm{, otherwise,}
    \end{cases}
    \label{eq:mc_label}
\end{equation}
with $\mathbbm{1}$ denoting the indicator function.

Unlike what is commonly seen in active learning settings, our goal is not to maximize the total entropy but rather to minimize it in order to ensure a high-quality pseudo-label set. We therefore redefine the self-training objective (Eq.~\ref{eq:st}) for stochastic pseudo-labeling as:
\begin{equation} \label{eq:uas}
  \min_{\omega, \hat{\bm{\mathrm{y}}}}
    \sum_{d=1}^{D}
    \left ( \vphantom{\sum_{r=1}^R}
        \mathcal{L}_{sem}(\bm{\hat y}, \bm{\mathrm{y}}) + (\mathbb{H}[y|x,D] - \tau) (\hat{\bm{\textrm{y}}}^* \cdot \hat{\bm{\mathrm{y}}})
    \right )
\end{equation}
with $\tau \in (0,1)$ determining the cut-off entropy threshold. To solve the nonlinear integer optimization task, we employ the following solver:
\begin{equation} \label{eq:pl_solver}
    \hat{\mathrm{y}}^{(c)} = 
    \begin{cases}
    1\textrm{, if } \mathbb{H}[y|x,D] < \tau \textrm{ and}\\
    \hphantom{1\textrm{, if }}c = \argmax \hat{\bm{\textrm{y}}}^* \\
    0\textrm{, otherwise,}
    \end{cases}
\end{equation}
where labels $\hat{\mathrm{y}}^{(c)} = 0 \ \forall \ c$ are ignored during training.

In practice, following precedent~\cite{zou2018unsupervised,unal2022scribble}, we set $\tau$ based on $p_\tau$, with $p_\tau$ determining the percentage of pseudo-labels to be sampled amongst the accumulated predicted labels.

In summary, we first determine a set of labels that are agreed upon by all stochastic forward passes, i.e.\ by forward passes with active dropout~\cite{srivastava2014dropout, gal2016dropout}. We then compute the Shannon entropy for all predictions, filtering those that exhibit high entropy. This surprisingly simple but effective approach allows us to reduce the incorrect targets that we introduce to our pseudo-label set and to avoid the overconfidence issue of na\"ive thresholding-based approaches.

\subsection{Semi-Supervised 3D Instance Segmentation}\label{sec:semi_inst}

\begin{algorithm}[t]
\SetKwInput{KwInput}{Input}
\SetKwInput{KwOutput}{Return}
\DontPrintSemicolon
\KwInput{Point cloud $P \in \mathbb{R}^{N \times 3}$, trained model with dropout $\omega$, number of passes $K$, threshold $\tau$}
seed\_output = $\omega(P)$ \ $\in \mathbb{R}^{M\times N}$ \;
$\bm{i}$ = seed\_output.copy() \;
seed\_mask = to\_mask($\bm{i}$) \ $\in \{0,1\}^{M\times N}$ \;
unanimous = seed\_mask.copy() \;
\For{$k$ in $1\textrm{:K}$} {
    output = $\omega_k(a(P))$ \ $\in \mathbb{R}^{M'\times N}$ \;
    mask = to\_mask(output) \ $\in \{0,1\}^{M'\times N}$ \;
    cost = -\,iou(seed\_mask, mask) \;
    row, col = LSA(cost) \; % \ $\in \{1,\,\dots,\,M'\}^M",\,\{1,\,\dots,\,M\}^M" \textrm{ , with } M" \leq M$
    unanimous[col] \&= mask[row] \;
    $\bm{i}$[col] += output[row]
}
% c\_0 = (K - integral) * log(K - integral) \;
% c\_1 = integral * log(integral) \;
% entropy = c\_0 + c\_1 \;
$\mathbb{H} = (K - \bm{i}) \cdot \log (K - \bm{i}) + \bm{i} \log \bm{i}$ \;
pseudo\_label = ($\mathbb{H}$ $< \tau$) \& unanimous \;
\KwOutput{pseudo\_label $\in \{0,1\}^{M\times N}$}
\caption{Bayesian pseudo-label generation for instance segmentation.}
\label{alg:st_inst}
\end{algorithm}

The proposed Bayesian self-training framework provides a strong basis for any semi-supervised 3D dense prediction task, one such example being 3D instance segmentation. Yet, compared to semantic segmentation where accumulation of labels can be easily handled across multiple stochastic forward passes, accumulation of predicted instance masks is not trivial and requires $n$-partite matching.

However, the $n$-partite matching problem is NP-hard with a complexity of $|I|!^{(K-1)}$ with $K$ passes and $|I|$ number of instances assuming full bijection. To match predicted instance masks across various stochastic forward passes, we propose a heuristic method based on the Hungarian algorithm. First, we generate an initial prediction via an unaugmented forward pass (Alg.~\ref{alg:st_inst}~L1\&3). The goal is to establish a seed that is likely of a higher quality with minimal perturbations of the data. We then iteratively assign predicted masks of each stochastic forward pass via linear sum assignment, with the cost matrix formed by the negative pairwise IoUs (Alg.~\ref{alg:st_inst}~L5-10).

Having matched predicted binary instance masks from individual stochastic forward passes, each resulting set of masks can then be treated as a set of 2-class semantic segmentation outputs, and thus our proposed Bayesian pseudo-labeling approach from Eq.~\ref{eq:pl_solver} can be trivially applied (Alg.~\ref{alg:st_inst}~L11-15). Our pipeline for semi-supervised 3D instance segmentation is illustrated in Fig.~\ref{fig:inst_pipeline}.

\subsection{Semi-Supervised Dense 3D Visual Grounding}
\label{sec:semi_vg}

Similar to both semi-supervised 3D semantic and 3D instance segmentation, our proposed Bayesian self-training formulation can easily be extended to include natural language prompts as inputs and thus tackle semi-supervised dense 3D visual grounding, i.e., referral-based 3D instance segmentation.

A common technique when tackling dense 3D visual grounding is to employ a grounding-by-selection strategy which can be summarised in two steps~\cite{chen2020scanrefer, zhao20213dvgt, unal2023three}:
\begin{enumerate}
    \item \textbf{Visual:} First, a 3D backbone computes object candidates $O$ from the scene.
    \item \textbf{Verbo-visual:} Then, a verbo-visual fusion module $\omega_{sel}$ selects the correct instance candidate from predicted instances $I$ based on the natural language description $W$, i.e., it predicts the index:
    \begin{equation}
        i^* = \argmax \, G \textrm{, with } G = \omega_{sel}(O, W) \in \mathbb{R}^{|I|},
    \end{equation}
\end{enumerate}
To extend our Bayesian self-training approach to this task, we follow the above structure and individually adapt each of the two steps, with the adaptation of the visual step simply following Sec.~\ref{sec:semi_inst}.

For the verbal side of 3D visual grounding, i.e.\ referral-based instance selection, we observe that the pseudo-label generation can be reduced to the same formulation as the pseudo-label generation for 3D instance segmentation. Utilizing the results from the heuristic $n$-partite matching (Alg.~\ref{alg:st_inst} L9), vector $G$ can be reordered to match across stochastic forward passes. Again, having matched predicted probability vectors, the method presented in Sec.~\ref{sec:semi_sem} can be trivially applied (Alg.~\ref{alg:st_inst}  L11-15).

\section{Experiments}

We evaluate our method on all three semi-supervised 3D semantic perception tasks: 3D semantic segmentation, 3D instance segmentation, and dense 3D visual grounding. The dataset and implementation details regarding each task can be found in the supplementary materials.

\subsection{3D Semantic Segmentation}\label{sec:exp_sem}

% \noindent \textbf{Implementation details:} For 3D semantic segmentation, we use Cylinder3D~\cite{} in our experiments without modification for a fair comparison to existing work~\cite{}. We use a labeling percentage of $p_\tau=0.75$ and only require a single iteration of self-training.

% \noindent \textbf{Dataset:} We run our experiments on the SemanticKITTI~\cite{} which is the most popular dataset for LiDAR semantic segmentation. SemanticKITTI consists of 11 sequences, with sequence 8 reserved for validation. We further show results on ScribbleKITTI~\cite{unal2022scribble}, which is a realistically weakly labeled dataset built on top of SemanticKITTI consisting of only $8\%$ labeled points. Following precedent we uniformly sample the data at varying thresholds~\cite{} and report the mIoU on the \textit{val}-set.

\begin{figure}[t]
    \centering
    \includegraphics[width=\textwidth]{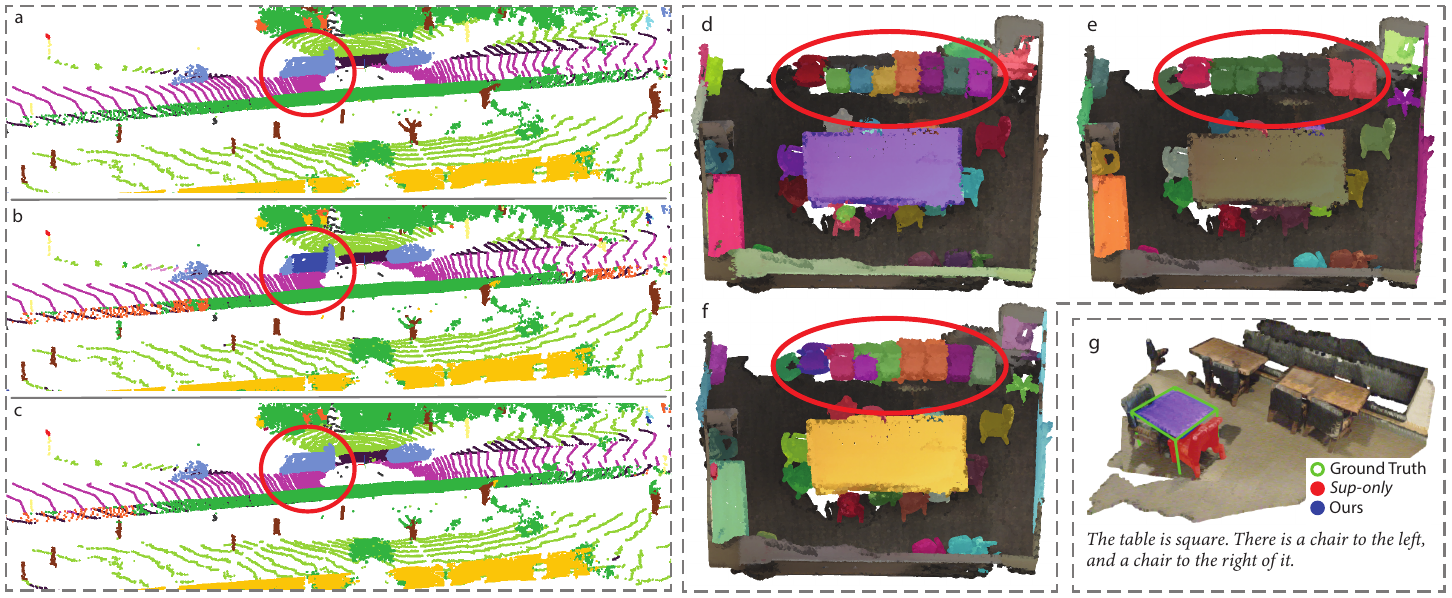}
    \caption{Qualitative results from the SemanticKITTI ($10\%$) \textit{val}-set, comparing a) the ground truth, b) \textit{Sup-only} and c) ours; on ScanNet ($10\%$), comparing d) the ground-truth instance masks, e) \textit{Sup-only} and f) ours; and finally g) ScanRefer ($10\%$). }
    \label{fig:result}
\vspace{-5px} \end{figure}

The comparison to the state of the art for semi-supervised 3D semantic segmentation is shown in Tab.~\ref{tab:results_kitti}. As seen, our Bayesian self-training strategy outperforms existing work in both SemanticKITTI~\cite{behley2019semantickitti} and ScribbleKITTI~\cite{unal2022scribble} on splits of $10\%$, $20\%$ and $50\%$, while performing on par for $1\%$. Specifically, compared to the baseline Cylinder3D~\cite{zhu2021cylindrical} model denoted as \textit{Sup-only}, our strategy allows significant improvements ranging from $4.4\%$ to $10.9\%$ mIoU. In Fig.~\ref{fig:result} (a-c) we showcase qualitative results on SemanticKITTI $10\%$.

\begin{table}[t]
    \centering
    \caption{Comparison of state-of-the-art methods for semi-supervised LiDAR semantic segmentation (mIoU [$\%$]). For fair comparison, all methods use the same backbone given by \textit{Sup-only}, which at $100\%$ labels achieves $64.3\%$ mIoU on SemanticKITTI and $57.0\%$ mIoU on ScribbleKITTI.} \vspace{-5px}
    \tabcolsep=0.27cm
    \begin{tabular}{|c|l|cccc|}
    \cline{2-6}
    \multicolumn{1}{c|}{} & Labeled & $1\%$ & $10\%$ & $20\%$ & $50\%$ \\
    \hline
    \multirow{8}{*}{\rotatebox[origin=c]{90}{SemanticKITTI~\cite{behley2019semantickitti}}} & \textit{Sup-only}~\cite{zhu2021cylindrical} & 45.4 & 56.1 & 57.8 & 58.7 \\
    \cline{2-6}
    & DiAL~\cite{unal2023discwise} & 45.4 & 57.1 & 59.2 & 60.0 \\
    & CBST~\cite{zou2018unsupervised} & 48.8 & 58.3 & 59.4 & 59.7 \\
    & CPS~\cite{chen2021semi} & 46.7 & 58.7 & 59.6 & 60.5 \\
    & LaserMix~\cite{kong2023lasermix} & \textbf{50.6} & 60.0 & 61.9 & 62.3 \\
    & IGNet~\cite{unal2024ignet} & 49.0 & 61.3 & 63.1 & \textbf{64.8} \\
    \cline{2-6}
    & Ours & 49.8 & \textbf{61.7} & \textbf{63.7} & 64.1 \\
    & $\Delta$ \textit{Sup-only} & \textbf{\color{fgreen}+4.4} & \textbf{\color{fgreen}+5.6} & \textbf{\color{fgreen}+5.9} & \textbf{\color{fgreen}+5.4} \\
    \hline
    \hline
    \multirow{8}{*}{\rotatebox[origin=c]{90}{ScribbleKITTI~\cite{unal2022scribble}}} & \textit{Sup-only}~\cite{zhu2021cylindrical} & 39.2 & 48.0 & 52.1 & 53.8 \\
    \cline{2-6}
    & DiAL~\cite{unal2023discwise} & 41.0 & 50.1 & 52.8 & 53.9 \\
    & CBST~\cite{zou2018unsupervised} & 41.5 & 50.6 & 53.3 & 54.5 \\
    & CPS~\cite{chen2021semi} & 41.4 & 51.8 & 53.9 & 54.8\\
    & LaserMix~\cite{kong2023lasermix} & 44.2 & 53.7 & 55.1 & 56.8\\
    & IGNet~\cite{unal2024ignet} & \textbf{44.4} & 57.7 & 59.6 & 60.8 \\
    \cline{2-6}
    & Ours & 43.9 & \textbf{58.9} & \textbf{60.1} & \textbf{61.2} \\
    & $\Delta$ \textit{Sup-only} & \textbf{\color{fgreen}+4.7} & \textbf{\color{fgreen}+10.9} & \textbf{\color{fgreen}+8.0} & \textbf{\color{fgreen}+7.4} \\
    \hline
    \end{tabular}
    \label{tab:results_kitti}
\vspace{-5px} \end{table}

\subsection{3D Instance Segmentation} \label{sec:exp_inst}

\begin{table}[t]
\tabcolsep=0.015cm
    \centering
    \caption{Comparison to state-of-the-art methods for semi-supervised 3D instance segmentation on the ScanNet Limited Reconstruction \textit{valid}-set. For a fair comparison, all methods apart from that marked with * use the same backbone~\cite{hou2021exploring, wu20223d} as \textit{Sup-only}.} \vspace{-5px}
    \begin{tabular}{|l|ccc|ccc|ccc|ccc|}
         \hline
         Labeled & \multicolumn{3}{c|}{$1\%$} & \multicolumn{3}{c|}{$5\%$} & \multicolumn{3}{c|}{$10\%$} & \multicolumn{3}{c|}{$20\%$} \\
         Metric & mAP & AP50 & AP25 & mAP & AP50 & AP25 & mAP & AP50 & AP25 & mAP & AP50 & AP25 \\
         \hline
         \textit{Sup-only}~\cite{chu2022twist} & 5.1 & 9.8 & 17.6 & 18.2 & 32.0 & 47.0 & 26.7 & 42.8 & 58.9 & 29.3 & 47.9 & 63.0 \\
         \hline
         PointCont.~\cite{xie2020pointcontrast} & 7.2 & 12.5 & 20.3 & 19.4 & 35.4 & 48.5 & 27.0 & 43.9 & 59.5 & 30.2 & 49.5 & 63.6  \\
         CSC~\cite{hou2021exploring} & 7.1 & 13.0 & 21.2 & 20.9 & 36.7 & 50.6 & 27.3 & 45.0 & 60.2 & 30.6 & 50.3 & 64.1 \\
         TWIST~~\cite{chu2022twist} & \textbf{9.6} & 17.1 & \textbf{26.2} & \textbf{27.0} & 44.1 & \textbf{56.2} & 30.6 & 49.7 & 63.0 & 32.8 & 52.9 & 66.8 \\
         WS3D*~\cite{liu2022weakly} & - & \textbf{32.5} & - & - & \textbf{45.6} & - & - & 49.2 & - & - & 51.3 & - \\
         \hline
         Ours & 7.2 & 14.2 & 23.0 & 24.2 & 39.3 & 51.4 & \textbf{32.7} & \textbf{51.9} & \textbf{65.5} & \textbf{37.4} & \textbf{56.1} & \textbf{68.9} \\
         $\Delta$ \textit{Sup-only} & \textbf{\color{fgreen}+2.1} & \textbf{\color{fgreen}+4.4} & \textbf{\color{fgreen}+5.4} & \textbf{\color{fgreen}+6.0} & \textbf{\color{fgreen}+7.3} & \textbf{\color{fgreen}+4.4} & \textbf{\color{fgreen}+6.0} & \textbf{\color{fgreen}+9.1} & \textbf{\color{fgreen}+6.6} & \textbf{\color{fgreen}+8.1} & \textbf{\color{fgreen}+8.2} & \textbf{\color{fgreen}+5.9}  \\
    \hline
    \end{tabular}
    \label{tab:results_scannet}
\vspace{-15px} \end{table}

\begin{table}[t]
\tabcolsep=0.3cm
    \centering
    \caption{Comparison to state-of-the-art methods for semi-supervised 3D instance segmentation on S3DIS Area 5 via AP50 [$\%$]. For a fair comparison, all methods use the same backbone Sparse U-Net~\cite{hou2021exploring, wu20223d} as \textit{Sup-only}.} \vspace{-5px}
    \begin{tabular}{|l|c|c|c|c|}
         \hline
         Labeled & $5\%$ & $10\%$ & $20\%$ & $50\%$ \\
         \hline
         \textit{Sup-only}~\cite{chu2022twist} & 30.4 & 36.8 & 41.2 & 46.5 \\
         \hline
         PointContrast~\cite{xie2020pointcontrast} & 33.6 & 38.7 & 43.1 & 48.9 \\
         CSC~\cite{hou2021exploring} & 34.2 & 41.0 & 44.7 & 50.4 \\
         TWIST~\cite{chu2022twist} & \textbf{37.1} & \textbf{45.6} & 48.4 & - \\
         \hline
         Ours & 35.0 & 42.9 & \textbf{51.1} & \textbf{56.6} \\
         $\Delta$ \textit{Sup-only} & \textbf{\color{fgreen}+4.6} & \textbf{\color{fgreen}+6.1} & \textbf{\color{fgreen}+9.9} & \textbf{\color{fgreen}+10.1} \\
    \hline
    \end{tabular}
    \label{tab:results_s3dis}
\vspace{-10px} \end{table}

% \noindent \textbf{Implementation details:} For 3D instance segmentation, we use the same backbone Sparse U-Net~\cite{hou2021exploring, wu20223d} in our experiments. We use a labeling percentage of $p_\tau=0.75$ for instance masks and again only require a single iteration of self-training.

% \noindent \textbf{Dataset:} We evaluate our method on the indoor 3D datasets ScanNet~\cite{dai2017scannet} and S3DIS~\cite{armeni20163d}. ScanNet consists of 1613 scenes, of which 312 are reserved for validation. For the training splits, we follow the limited reconstruction setting from Hou~\etal~\cite{} and report the mAP and AP at threshold $50\%$ and $25\%$ on the \textit{val}-set. S3DIS is a much smaller dataset consisting of only 271 scenes collected from 6 different areas, of which we reserve Area 5 as the validation set following precedent~\cite{}. Due to its small size, we tackle the semi-supervised setting for labeling splits ranging from $5\%$ up to $50\%$.

For semi-supervised 3D instance segmentation, we report results for ScanNet~\cite{dai2017scannet} in Tab.~\ref{tab:results_scannet} and for S3DIS~\cite{armeni20163d} in Tab.~\ref{tab:results_s3dis}. As seen, we observe a similar behavior to semi-supervised 3D semantic segmentation, where at higher thresholds of $10\%$, $20\%$ for ScanNet and $20\%$, $50\%$ for S3DIS, our simple yet effective Bayesian self-training strategy significantly outperforms state-of-the-art methods that require highly engineered solutions. In contrast, at lower thresholds, we see a significant gap in performance (especially for ScanNet). Still, compared to the baseline \textit{Sup-only}, we see significant gains across the board for all semi-splits. Additionally in Fig.~\ref{fig:result} (d-f), we show qualitative results on the ScanNet \textit{val}-set.

\subsection{Dense 3D Visual Grounding}

% \noindent \textbf{Implementation details:} For 3D visual grounding, we use the current state-of-the-art ConcreteNet~\cite{unal2023three} in our experiments. We use $p_\tau=75\%$ for the grounding pseudo-labels when including verbal cues in the unlabeled set and only require a single iteration of self-training.

% \noindent \textbf{Dataset:} We evaluate our method on the ScanRefer~\cite{} dataset using axis-aligned bounding boxes that are fitted onto the predicted instance masks. ScanRefer builds on top of ScanNet but only consist of ?? scenes, of which ?? are reserved for validation. We construct three semi-supervised splits of $5\%$, $10\%$ and $20\%$ where we ensure the data split is valid for both the number of 3D scenes and the number of available utterances.

For semi-supervised 3D visual grounding, we report results on the ScanRefer~\cite{chen2020scanrefer} dataset. As seen in Tab.~\ref{tab:results_scanrefer} (Vi), our self-training approach shows improvements across the board, with significant gains on the ``unique'' category.

While natural language prompts still require human effort to generate, they are considerably cheaper and easier to acquire than per-point high-quality panoptic labels. In Tab.~\ref{tab:results_scanrefer} (ViVe), we show that the inclusion of such prompts in the unlabeled dataset further benefits the model's accuracy. As seen, we observe even larger gains over the baseline method with the overall accuracy (OA) showing $+4.7\%$ improvement with a $50\%$ IoU threshold.

Furthermore, in Fig.~\ref{fig:result} (g) we qualitatively compare our method to the baseline \textit{Sup-only} on ScanRefer.

% \begin{figure}[t]
%     \centering
%     \includegraphics[width=\textwidth]{figures/result_instance.pdf}
%     \caption{Qualitative results on ScanNet ($10\%$), comparing a) the ground-truth instance masks with b) \textit{Sup-only} and c) ours. In d) we show results on ScanRefer ($10\%$).}
%     \label{fig:result}
% \vspace{-15px} \end{figure}

\begin{table}[t]
    \centering
    \caption{Results for semi-supervised 3D dense visual grounding while assuming (Vi) that the model is given no natural language descriptions for the unlabeled split and (ViVe) when the model has access to natural language descriptions for the unlabeled split. Reported values are given as Acc@25/50 for the 10\% split.} \vspace{-5px}
    \begin{tabular}{|l|l|ccc|}
        \cline{2-5}
        \multicolumn{1}{c|}{} & Category & Unique & Multiple & Overall \\
        \cline{2-5}
        \multicolumn{1}{c|}{} & \textit{Sup-only}~\cite{unal2023three} & 51.9 / 43.6 & 19.3 / 14.1 & 25.4 / 19.6 \\
        \hline
        \multirow{2}{*}{\rotatebox[origin=c]{90}{Vi}} &
         Ours & 58.8 / 48.8 & 22.0 / 18.2 & 28.8 / 23.9 \\
         & $\Delta$ \textit{Sup-only} & \textbf{\color{fgreen}+6.9} / \textbf{\color{fgreen}+5.2} & \textbf{\color{fgreen}+2.7} / \textbf{\color{fgreen}+4.1} & \textbf{\color{fgreen}+3.4} / \textbf{\color{fgreen}+4.3} \\
         \hline
        \multirow{2}{*}{\rotatebox[origin=c]{90}{ViVe}} & Ours & 61.0 / 51.2 & 23.4 / 18.1 & 30.4 / 24.3 \\
        & $\Delta$ \textit{Sup-only} & \textbf{\color{fgreen}+9.1} / \textbf{\color{fgreen}+7.6} & \textbf{\color{fgreen}+4.1} / \textbf{\color{fgreen}+4.0} & \textbf{\color{fgreen}+5.0} / \textbf{\color{fgreen}+4.7} \\
        \hline
    \end{tabular}
    \label{tab:results_scanrefer}
\vspace{-5px} \end{table}

\begin{table}[t]
    \centering
    \caption{Comparison of the pseudo-label quality of current thresholding strategies and their effects on performance on SemanticKITTI 10\%, ScanNet 10\%, ScanRefer 10\% (with visual and verbal) for each respective task. Pseudo-label thresholding is applied after unanimous voting for a fair comparison. The letters indicate each pseudo-label type: $S$- for semantic, $I$- for instance, and $G$- for grounding.} \vspace{-5px}
    \tabcolsep=0.065cm
    \begin{tabular}{|l|cc|ccc|cccc|}
        \cline{2-10}
        \multicolumn{1}{c|}{} & \multicolumn{2}{c|}{Semantic} & \multicolumn{3}{c|}{Instance} & \multicolumn{4}{c|}{Visual Grounding} \\
        \hline
        Method & $S$-Acc & mIoU & $S$-Acc & $I$-Acc & AP@50 & $S$-Acc & $I$-Acc & $G$-Acc & OA@50 \\
        \hline
        \textit{Sup-only~\cite{unal2023three}} & - & 56.1 & - & - & 42.8 & - & - & - & 19.6 \\
        \hline
        \textit{Unanimous} & 96.4 & 60.0 & 79.3 & 75.3 & 47.0 & 72.9 & 69.9 & 61.5 & 16.9 \\
        \hline
        Na\"ive~\cite{cascante2021curriculum, xie2020selftraining, zoph2020rethinking} & 98.3 & 60.8 & 82.9 & 77.7 & 50.2 & 82.0 & 73.0 & 68.6 & 21.9 \\
        Class-balanced~\cite{zou2018unsupervised} & 96.6 & 61.1 & 81.2 & 70.1 & 44.4 & 80.9 & 69.0 & 60.6 & 16.2 \\
        \hline
        Entropy-based & \textbf{98.6} & \textbf{61.7} & \textbf{84.8} & \textbf{77.8} & \textbf{51.9} & \textbf{83.8} & \textbf{75.2} & \textbf{75.6} & \textbf{24.3} \\ 
        $\Delta$ \textit{Unanimous} & \textbf{\color{fgreen}+2.2} & \textbf{\color{fgreen}+1.7} & \textbf{\color{fgreen}+5.5} & \textbf{\color{fgreen}+2.5} & \textbf{\color{fgreen}+4.9} & \textbf{\color{fgreen}+5.8} & \textbf{\color{fgreen}+2.3} & \textbf{\color{fgreen}+5.7} &\textbf{\color{fgreen}+7.4} \\
        \hline
    \end{tabular}
    \label{tab:ablation_thresholding}
\vspace{-0px} \end{table}

\begin{figure}[t]
    \centering
    \includegraphics[width=\columnwidth]{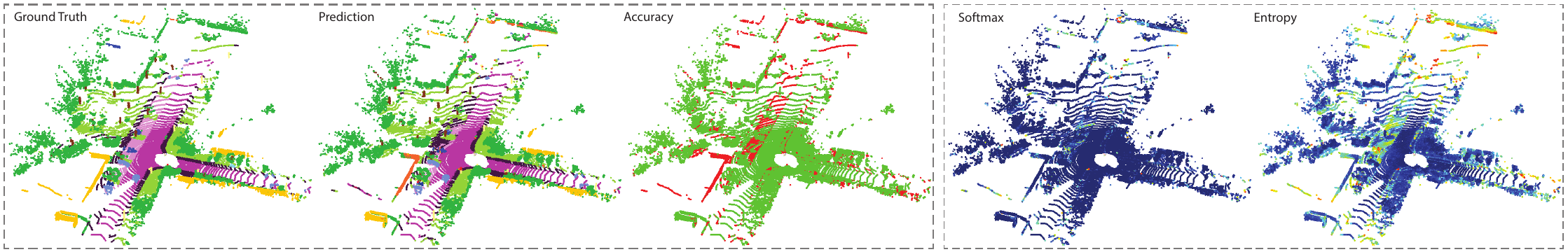}
    \caption{Qualitative analysis of the uncertainty estimation. We compare softmax confidence to our MC-derived Shannon entropy.}
    \label{fig:rebuttal}
\vspace{-10px} \end{figure}

\begin{table}[t]
    \centering
    \caption{Ablation study analyzing the effect of the hyperparameter $p_\tau$ for each task. We vary the labeling percentage and show the performance on SemanticKITTI 10\%, ScanNet 10\%, and ScanRefer 10\% (incl.\ verbal input) for each respective task.} \vspace{-5px}
    \tabcolsep=0.45cm
    \begin{tabular}{|l|c|c|c|}
        \hline
        Threshold & mIoU & AP@50 & OA@50 \\
        \hline
        $50\%$ & 61.4 & 50.0 & 22.8 \\
        $75\%$ & \textbf{61.7} & \textbf{51.9} & \textbf{24.3} \\
        $100\%$ & 60.0 & 47.0 & 16.9 \\
        \hline
    \end{tabular}
    \label{tab:ablation_labeling_percentage}
\vspace{-5px}\end{table}

\begin{table}[t]
    \centering
    \caption{Ablation study showing the performance on SemanticKITTI 10\%, ScanNet 10\% and ScanRefer 10\% (incl.\ verbal input) for varying label aggregation strategies from majority rules to unanimous voting over 9 stochastic forward passes.} \vspace{-5px}
    \tabcolsep=0.45cm
    \begin{tabular}{|l|c|c|c|}
        \hline
        Threshold & mIoU & AP@50 & OA@50 \\
        \hline
        5/9 & 60.7 & 47.9 & 20.1 \\
        7/9 & 61.0 & 51.0 & 23.9 \\
        9/9 & \textbf{61.7} & \textbf{51.9} & \textbf{24.3} \\
        \hline
    \end{tabular}
    \label{tab:ablation_unanimous}
\vspace{-10px} \end{table}

\subsection{Ablation Studies}

We conduct ablation studies on $10\%$ splits of SemanticKITTI~\cite{behley2019semantickitti}, ScanNet~\cite{dai2017scannet} and ScanRefer~\cite{chen2020scanrefer} for semi-supervised 3D semantic segmentation, instance segmentation and visual grounding respectively.

\noindent\textbf{Entropy Thresholding:} In Tab.~\ref{tab:ablation_thresholding}, we show an analysis of the effectiveness of our proposed uncertainty thresholding via Shannon entropy, comparing it to the baseline unanimous voting results as well as established thresholding techniques of softmax thresholding~\cite{xie2020selftraining} and class-balanced thresholding~\cite{zou2018unsupervised}. Alongside the performance for each individual task, we also report the accuracy of the generated pseudo-labels. For instance segmentation, a pseudo-mask is considered correct if its intersection with the associated ground-truth mask is identical to the former. For visual grounding, we reduce the accuracy threshold to $0.5$ following the benchmark metric used. As seen, our uncertainty-based filtering improves pseudo-label quality across the board. We also showcase the performance implications of uncertainty filtering, where it can be observed that entropy thresholding outperforms both unanimous voting and current thresholding strategies across all tasks. {We further provide a qualitative analysis of the uncertainty metric in Fig.~\ref{fig:rebuttal} to illustrate the better correlation of our MC-derived uncertainty with pseudo-label accuracy compared to softmax based uncertainty.}

\noindent\textbf{Labeling Percentage ($p_\tau$):} We compare various set labeling percentages $p_\tau$ across all tasks in Tab.~\ref{tab:ablation_labeling_percentage}. While we obtain higher accuracy pseudo-labels at $50\%$, it can be seen that the reduced supervision negatively impacts performance.

\section{Limitations and Discussion}

The main advantage of the proposed Bayesian self-training is its simplicity and ease of implementation compared to existing SOTA approaches that utilize highly engineered and often costly modules. However, the method has a limitation. As observed from the results, the performance is comparably inferior to SOTA semi-supervised approaches for highly sparse training splits. This behavior stems from the reliance on unanimous voting (Eq.~\ref{eq:mc_label}). With low amounts of labeled data, the initial na\"ive model performance suffers, causing a reduction in the number of unanimously voted points. This effect is further pronounced when tackling instance segmentation, where fused instance masks may affect multiple predictions. In Tab.~\ref{tab:ablation_unanimous}, we show that relaxation does not aid the performance as it further introduces incorrect predictions into the pseudo-label set.

While semi-supervised training methods that enhance the robustness of the pseudo-labels either via a teacher model (DiAL~\cite{tarvainen2017mean, unal2023discwise}), mix-augmentations (LaserMix~\cite{kong2023lasermix}) or multi-stage refinement (TWIST~\cite{chu2022twist}) are presented as competing to ours, we argue that as a solution, they can in fact be utilized within our framework in place of the 3D model to improve the quality of the initial predictions. This would allow a larger set of unanimously voted samples and would thus re-enable the benefits of our proposed method on highly sparse splits.

\section{Conclusion}

In this work, we introduce a {simple yet effective} Bayesian self-training approach for semi-supervised 3D semantic segmentation {by utilizing uncertainty estimation strategies from the field of active learning. Specifically,} through multiple stochastic forward passes, we generate an initial pseudo-label set and then filter it based on estimated predictive {entropy}. We further extend our {semi-supervised 3D} method to {newly} include both instance segmentation and referral-based instance segmentation via an {efficient,} heuristic $n$-partite matching algorithm. We conduct extensive experiments on multiple datasets, covering both indoor and outdoor domains, {showing that our method performs on par if not better than current domain- and task-specific methods.}

\vspace{5px}
\noindent\textbf{Acknowledgments:} This work is funded by Toyota Motor Europe via the research project TRACE-Z\"urich.

\clearpage
\clearpage

\bibliographystyle{splncs04}
\bibliography{main}

\clearpage
\clearpage

\appendix
%%%%%%%%%%%%%%%%%%%%%%%%%%%%%%%%%%%%%%%%%%%%%%%%%%%%%%%%%%%%%%%%%%%%%%%%%%%%%%%%%%%%%%%%%%%%%%%%%%%%%%%%%%%%%%%%%%%%%%%%%%%%%%%%%%%%%%%%%%%%%%%%%%%%%%%%%%%%%%%%%%%%%%%%%%%%%%%%%%%%%%%%%%%%%%%%%%%%%%%%%%%%%%%%%%%%%%%%%%%%%%%%%%%%%%%%%%%%%%%%%%%%%%%%%%%%%%%%%%%%%%%%%%%%%%%%%%%%%%%%%%%%%%%%%%%%%%%%%%%%%%%%%%%%%%%%%%%%%%%%%%%%%%%%%%%%%%%%%%%%%%%%%%%%%%%%%%%%%%%%%%%%%%%%%%%%%%%

%%%%%%%%%%%%%%%%%%%%%%%%%%%%%%%%%%%%%%%%%%%%%%%%%%%%%%%%%%%%%%%%%%%%%%%%%%%%%%%%%%%%%%%%%%%%%%%%%%%%%%%%%%%%%%%%%%%%%%%%%%%%%%%%%%%%%%%%%%%%%%%%%%%%%%%%%%%%%%%%%%%%%%%%%%%%%%%%%%%%%%%%%%%%%%%%%%%%%%%%%%%%%%%%%%%%%%%%%%%%%%%%%%%%%%%%%%%%%%%%%%%%%%%%%%%%%%%%%%%%%%%%%%%%%%%%%%%%%%%%%%%%%%%%%%%%%%%%%%%%%%%%%%%%%%%%%%%%%%%%%%%%%%%%%%%%%%%%%%%%%%%%%%%%%%%%%%%%%%%%%%%%%%%%%%%%%%%

\section{Implementation Details and Dataset}

\subsection{3D Semantic Segmentation}

\noindent \textbf{Implementation details:} For 3D semantic segmentation, we use Cylinder3D~\cite{zhu2021cylindrical} in our experiments without modification for a fair comparison to existing work. We use a labeling percentage of $p_\tau=0.75$ and only require a single iteration of self-training. We use K=9 passes.

\noindent \textbf{Dataset:} We run our experiments on the SemanticKITTI~\cite{behley2019semantickitti} which is the most popular dataset for LiDAR semantic segmentation. SemanticKITTI consists of 11 sequences, with sequence 8 reserved for validation. We further show results on ScribbleKITTI~\cite{unal2022scribble}, a realistically weakly labeled dataset built on top of SemanticKITTI consisting of only $8\%$ labeled points. Following precedent we uniformly sample the data at varying thresholds~\cite{kong2023lasermix} and report the mIoU on the \textit{val}-set.

\subsection{3D Instance Segmentation} 

\noindent \textbf{Implementation details:} For 3D instance segmentation, we use the same backbone Sparse U-Net~\cite{hou2021exploring, wu20223d} in our experiments. We use a labeling percentage of $p_\tau=0.75$ for instance masks and again only require a single iteration of self-training. We use K=9 passes.

\noindent \textbf{Dataset:} We evaluate our method on the indoor 3D datasets ScanNet~\cite{dai2017scannet} and S3DIS~\cite{armeni20163d}. ScanNet consists of 1613 scenes, of which 312 are reserved for validation. For the training splits, we follow the limited reconstruction setting from Hou~\etal~\cite{hou2021exploring} and report the mAP and AP at threshold $50\%$ and $25\%$ on the \textit{val}-set. S3DIS is a much smaller dataset consisting of only 271 scenes collected from 6 different areas, of which we reserve Area 5 as the validation set following precedent~\cite{chu2022twist}. Due to its small size, we tackle the semi-supervised setting for labeling splits ranging from $5\%$ up to $50\%$.

\subsection{Dense 3D Visual Grounding}

\noindent \textbf{Implementation details:} For 3D visual grounding, we use the current state-of-the-art ConcreteNet~\cite{unal2023three} in our experiments. We use $p_\tau=75\%$ for the grounding pseudo-labels when including verbal cues in the unlabeled set and only require a single iteration of self-training. We use K=9 passes.

\noindent \textbf{Dataset:} We evaluate our method on the ScanRefer~\cite{chen2020scanrefer} dataset using axis-aligned bounding boxes that are fitted onto the predicted instance masks. ScanRefer builds on top of ScanNet but only consists of a single furniture arrangement per room. We construct three semi-supervised splits of $5\%$, $10\%$, and $20\%$ where we ensure the data split is valid for both the number of 3D scenes and the number of available utterances.

\subsection{Classwise 3D Semantic Segmentation Results}

In reference to the Tab.~1 on the main manuscript, we provide Tab.~\ref{tab:supp_semantickitti} and Tab.~\ref{tab:supp_scribblekitti} that show the classwise IoUs for the same semi-supervised 3D semantic segmentation experiments on SemanticKITTI and ScribbleKITTI respectively (apart from \cite{unal2024ignet}). Here we observe two key feats of our Bayesian self-training method: (i) Our method outperforms existing work in head classes that are more prone to the softmax overconfidence issue; (ii) our method shows great improvements over the baseline on classes that have similar 3D shapes (e.g. car, truck, other vehicle). The inclusion of uncertainty-based filtering allows the network to reduce the number of false positives, allowing better separation between geometrically similar objects.

\begin{table*}[t]
    \caption{Classwise results for semi-supervised 3D semantic segmentation on SemanticKITTI~\cite{behley2019semantickitti}.
    \label{tab:supp_semantickitti}} \vspace{-5px}
    \resizebox{\textwidth}{!}{
    \begin{tabular}{|l|l|c|ccccccccccccccccccc|}
        \cline{2-22}
        \multicolumn{1}{c|}{} & Method
        & mIoU
        &\lturn{car}
        &\lturn{bicycle}
        &\lturn{m.cycle}
        &\lturn{truck} 
        &\lturn{o.vehicle } 
        &\lturn{person}
        &\lturn{bicyclist} 
        &\lturn{m.cyclist} 
        &\lturn{road}
        &\lturn{parking}  
        &\lturn{sidewalk} 
        &\lturn{o.ground} 
        &\lturn{building} 
        &\lturn{fence}
        &\lturn{vegetation } 
        &\lturn{trunk}
        &\lturn{terrain} 
        &\lturn{pole}
        &\lturn{t.sign} \\
        [0.5ex] 
        \hline
        \multirow{6}{*}{\rotatebox[origin=c]{90}{$1\%$}} &\textit{Sup-only}~\cite{zhu2021cylindrical} & 45.4 & 90.9 & 24.5 & 2.8 & 35.1 & 20.4 & 31.7 & 49.5 & 0.0 & 85.5 & 23.4 & 67.5 & \textbf{1.3} & 85.0 & 46.0 & 84.1 & 49.1 & 70.3 & 55.0 & 40.6
\\
        \cline{2-22}
        &DiAL~\cite{unal2023discwise} & 45.4&  91.2&  13.2&  5.4 & 47.3&  14.5&  29.0 & 37.3 & 0.0 & 86.8 & 22.6 & 70.3&  1.2&  86.7 & 45.4&  84.7&  59.4 & 70.9&  55.8& 40.8
\\
        &CBST~\cite{zou2018unsupervised} & 48.8 & 92.4 & 16.3 & 6.4&  \textbf{61.9}&  \textbf{27.0} & 35.7 & 49.4&  0.0 & 88.9 & 29.4 & 73.2 & 0.7 & \textbf{89.1} & 49.5 & 83.9&  51.4 & 68.1&  59.8&  44.0
 \\
        &CBST~\cite{chen2021semi} & 46.7 & 92.0 & 13.5&  7.1&  37.8&  12.7 & 33.0&  \textbf{54.5} & 0.0 & 89.8 & 25.0 & 73.8 & 0.0 & 88.8 & \textbf{50.1} & 83.6 & 57.4 & 67.8 & 58.2 & 42.1
\\
        &LaserMix~\cite{kong2023lasermix} & \textbf{50.6} & 91.8 & \textbf{35.7} & 19.8 & 37.5 & 25.6 & \textbf{53.6} & 45.7 & \textbf{2.5} & 87.8 & \textbf{33.5} & 71.3 & 0.7 & 87.3 & 43.8 & 84.6 & \textbf{62.7} & 69.3 & 59.8 & \textbf{47.6}
\\
        \cline{2-22}
        &Ours & 49.8 &
\textbf{93.1} & 
21.3 & 
\textbf{24.8} & 
44.7 & 
20.8 & 
48.3 & 
38.3 & 
0.0 & 
\textbf{90.3} & 
29.5 & 
\textbf{73.9} & 
0.0 & 
85.0 & 
42.5 & 
\textbf{88.2} & 
59.5 & 
\textbf{75.1} & 
\textbf{64.6} & 
46.5\\
        \hline
        \hline
        \multirow{6}{*}{\rotatebox[origin=c]{90}{$10\%$}} &\textit{Sup-only}~\cite{zhu2021cylindrical} & 56.1 & 93.4 & 38.4 & 47.7 & 65.7 & 31.0 & 61.9 & 64.9 & 0.0 & 90.7 & 37.7 & 75.3 & 0.9 & 89.2 & 50.5 & 86.4 & 56.0 & 73.9 & 56.2 & 46.0
 \\
        \cline{2-22}
        &DiAL~\cite{unal2023discwise} & 57.1 & 94.1 & 40.5 & \textbf{58.4} & 56.0 & 38.0 & \textbf{66.5} & 75.6 & 0.0 & 88.4 & 22.7 & 72.0 & 1.5 & 87.9 & 49.3 & 86.7 & 66.1 & 74.2 & 58.0 & 49.2
\\
        &CBST~\cite{zou2018unsupervised} & 58.3 & 93.6 & 40.3 & 43.5 & 80.4 & 33.8 & 57.6 & 78.1 & 0.0 & 91.6 & 36.3 & 76.6 & \textbf{5.1} & 89.2 & 51.1 & 86.3 & 61.9 & 71.2 & 61.3 & 49.7
 \\
        &CBST~\cite{chen2021semi} & 58.7 & 94.0 & 38.7 & 51.0 & 60.3 & 39.8 & 65.7 & 80.0 & 0.0 & 91.4 & 33.2 & 76.4 & 2.9 & 89.8 & 53.8 & 87.2 & 65.7 & \textbf{74.6} & 61.5 & 50.0
\\
        &LaserMix~\cite{kong2023lasermix} & 60.0 & 93.8 & 44.9 & \textbf{58.4} & 65.6 & 39.4 & 65.8 & 80.9 & \textbf{0.2 }& 92.0 & \textbf{44.2} & 77.1 & 3.9 & 89.1 & 49.0 & 86.2 & \textbf{66.8} & 72.3 & 58.4 & 51.2
\\
        \cline{2-22}
        &Ours & \textbf{61.7} & \textbf{95.4} & \textbf{47.3} & 53.8 & \textbf{82.3} & \textbf{43.2} & 63.6 & \textbf{81.2} & 0.0 & \textbf{93.8} & 43.2 & \textbf{79.8} & 1.4 & \textbf{89.9} & \textbf{54.6} & \textbf{87.3} & 66.1 & 73.3 & \textbf{63.9} & \textbf{51.8} \\
        \hline
        \hline
        \multirow{6}{*}{\rotatebox[origin=c]{90}{$20\%$}} &\textit{Sup-only}~\cite{zhu2021cylindrical} & 57.8 & 94.0 & 31.6 & 47.3 & \textbf{89.5 }& 38.3 & 57.9 & 79.1 & 0.0 & 91.6 & 29.6 & 76.1 & 0.9 & 87.8 & 43.6 & 86.6 & 63.7 & 72.5 & 61.8 & 47.5
\\
        \cline{2-22}
        &DiAL~\cite{unal2023discwise} & 59.2 & 94.4 & 38.7 & 52.5 & 81.2 & 45.8 & 64.2 & 78.0 & 0.0 & 90.9 & 35.2 & 75.7 & 1.8 & 89.2 & 49.8 & 86.3 & 65.6 & 72.6 & 56.0 & 47.6
\\
        &CBST~\cite{zou2018unsupervised} & 59.4 & 94.2 & 41.8 & 51.4 & 77.7 & 39.8 & 65.4 & 79.8 & 0.0 & 91.7 & 29.8 & 76.3 & 3.5 & 89.2 & 49.7 & 87.1 & 66.1 & \textbf{74.2} & 60.1 & 51.3
\\
        &CBST~\cite{chen2021semi} & 59.6 & 94.2 & 41.8 & 52.9 & 78.2 & 39.6 & 66.1 & 80.6 & 0.0 & 91.9 & 30.2 & 76.4 & \textbf{3.7 }& 89.2 & 50.0 & 87.0 & 66.6 & 73.7 & 60.0 & 51.1
\\
        &LaserMix~\cite{kong2023lasermix} & 61.9 & 94.4 & 46.0 & \textbf{68.0} & 74.3 & 47.6 & 68.1 & 83.7 & 0.2 & 92.6 & \textbf{42.7} & 78.0 & 1.9 & 89.7 & 52.9 & 86.0 & \textbf{69.3} & 70.6 & 59.2 & 51.7
 \\
        \cline{2-22}
        &Ours & \textbf{63.7} &
\textbf{96.4} & 
\textbf{50.7} & 
61.6 & 
79.2 & 
\textbf{55.4} & 
\textbf{73.3} & 
\textbf{85.3} & 
\textbf{1.1} & 
\textbf{93.8} & 
40.6 & 
\textbf{80.1} & 
2.1 & 
\textbf{89.9} & 
\textbf{58.0} & 
\textbf{87.5} & 
66.4 & 
72.5 & 
\textbf{63.4} & 
\textbf{52.9}
\\
        \hline
        \hline
        \multirow{6}{*}{\rotatebox[origin=c]{90}{$50\%$}} &\textit{Sup-only}~\cite{zhu2021cylindrical} & 58.7 & 93.9 & 40.4 & 48.0 & 81.4 & 33.7 & 65.7 & 79.7 & 0.0 & 91.9 & 32.6 & 76.7 & 1.3 & 89.0 & 51.8 & 87.2 & 61.4 & 72.5 & 58.7 & 48.7
\\
        \cline{2-22}
        &DiAL~\cite{unal2023discwise} & 60.0 & 94.1 & 41.3 & 57.7 & 64.6 & 39.5 & 65.3 & 86.8 & 0.0 & 91.3 & 32.8 & 75.2 & 3.5 & 89.7 & 48.6 & 85.4 & 65.9 & 70.6 & 58.7 & 49.1
\\
        &CBST~\cite{zou2018unsupervised} & 59.7 & 94.9 & 40.9 & 54.4 & 75.3 & 43.8 & 67.3 & 86.8 & 0.0 & 91.5 & 33.3 & 75.7 & 2.6 & 89.3 & 50.7 & 86.7 & 63.9 & 72.4 & 56.4 & 48.8
 \\
        &CBST~\cite{chen2021semi} & 60.5 & 94.6 & 43.3 & 55.3 & 80.5 & 42.5 & 67.9 & 84.6 & 0.0 & 92.0 & 34.3 & 76.9 & 2.2 & 89.8 & 52.3 & 86.0 & 67.4 & 71.1 & 59.5 & 49.4
\\
        &LaserMix~\cite{kong2023lasermix} & 62.3 & 94.7 & 48.4 & \textbf{64.7} & 65.2 & 44.5 & 71.0 & \textbf{88.3} & \textbf{2.1} &  92.7 & \textbf{43.0} & 78.4 & 2.0 & \textbf{90.3} & \textbf{54.9} & \textbf{88.1} & \textbf{68.1} & \textbf{75.3} & \textbf{66.6} & 51.7
\\
        \cline{2-22}
        &Ours & \textbf{64.1} &
\textbf{96.4} & 
\textbf{51.1} & 
64.3 & 
\textbf{84.6} & 
\textbf{57.5} & 
\textbf{70.8} & 
83.9 & 
0.0 & 
\textbf{93.4} & 
41.5 & 
\textbf{79.6} & 
\textbf{6.3} & 
88.8 & 
53.8 & 
87.9 & 
67.8 & 
73.8 & 
64.3 & 
\textbf{52.7} \\
        \hline
    \end{tabular}
    }
\end{table*}

\begin{table*}[t]
    \tabcolsep=0.11cm
    \caption{Classwise results for semi-supervised 3D semantic segmentation on ScribbleKITTI~\cite{unal2022scribble}.
    \label{tab:supp_scribblekitti}} \vspace{-5px}
    \resizebox{\textwidth}{!}{
    \begin{tabular}{|l|l|c|ccccccccccccccccccc|}
        \cline{2-22}
        \multicolumn{1}{c|}{} & Method
        & mIoU
        &\lturn{car}
        &\lturn{bicycle}
        &\lturn{m.cycle}
        &\lturn{truck} 
        &\lturn{o.vehicle } 
        &\lturn{person}
        &\lturn{bicyclist} 
        &\lturn{m.cyclist} 
        &\lturn{road}
        &\lturn{parking}  
        &\lturn{sidewalk} 
        &\lturn{o.ground} 
        &\lturn{building} 
        &\lturn{fence}
        &\lturn{vegetation } 
        &\lturn{trunk}
        &\lturn{terrain} 
        &\lturn{pole}
        &\lturn{t.sign} \\
        [0.5ex] 
        \hline
        \multirow{6}{*}{\rotatebox[origin=c]{90}{$1\%$}} &\textit{Sup-only}~\cite{zhu2021cylindrical} & 39.2 & 83.2 & 13.8 & 3.4 & 26.3 & 11.8 & 28.0 & 25.2 & 0.0 & 72.5 & 13.0 & 59.5 & 0.2 & 86.6 & 33.7 & 78.7 & 55.7 & 58.4 & 54.0 & 40.3
\\
        \cline{2-22}
        &DiAL~\cite{unal2023discwise} & 41.0 & 82.3 & 15.8 & 7.1 & \textbf{32.0} & 15.4 & 23.7 & 36.3 & 0.0 & 75.0 & 12.6 & 61.4 & \textbf{0.9} & 85.3 & 30.0 & 80.1 & 57.0 & 67.0 & 56.1 & 41.3
\\
        &CBST~\cite{zou2018unsupervised} & 41.5 & 83.7 & 22.1 & 5.9 & 28.3 & 13.4 & 27.1 & 34.7 & 0.0 & 74.0 & 14.4 & 61.7 & 0.2 & \textbf{88.1} & 36.6 & 80.3 & 58.7 & 60.4 & 57.1 & 41.4
 \\
        &CBST~\cite{chen2021semi} & 41.4 & 82.8 & 18.2 & 11.4 & 20.9 & 15.1 & 22.5 & 35.5 & 0.0 & 74.7 & \textbf{15.7} & 61.6 & 0.4 & 86.0 & 34.2 & 82.2 & 58.4 & 69.9 & 56.7 & 40.0
\\
        &LaserMix~\cite{kong2023lasermix} & 44.2 & 82.6 & \textbf{25.5} & 18.8 & 29.0 & \textbf{19.8} & 41.1 & \textbf{47.2} & \textbf{0.6} & 71.5 & 10.5 & 64.2 & 2.2 & 85.1 & 33.5 & 82.0 & 59.9 & 65.8 & 54.5 & 45.2 
\\
        \cline{2-22}
        &Ours & \textbf{43.9} & \textbf{85.7} & 13.8 & \textbf{23.6} & 
7.9 & 
7.6 & 
\textbf{48.8} & 
30.2 & 
0.0 & 
\textbf{81.3} & 
\textbf{15.7} & 
\textbf{68.8} & 
0.2 & 
84.2 & 
\textbf{40.0} & 
\textbf{85.7} & 
\textbf{60.3} & 
\textbf{71.2} & 
\textbf{63.9} & 
\textbf{45.3} \\
        \hline
        \hline
        \multirow{6}{*}{\rotatebox[origin=c]{90}{$10\%$}} &\textit{Sup-only}~\cite{zhu2021cylindrical} & 48.0 & 85.7 & 25.6 & 21.3 & 52.8 & 29.9 & 46.5 & 47.2 & 0.1 & 79.5 & 15.4 & 63.8 & 0.3 & 85.4 & 39.6 & 84.8 & 59.7 & 71.5 & 57.7 & 45.8
\\
        \cline{2-22}
        &DiAL~\cite{unal2023discwise} & 50.1 & 83.7 & 32.6 & 45.1 & 41.0 & 34.7 & 56.0 & 59.2 & 0.0 & 75.9 & 14.0 & 64.0 & 0.7 & 85.6 & 37.9 & 83.3 & 62.6 & 68.2 & 59.7 & 47.0 
\\
        &CBST~\cite{zou2018unsupervised} & 50.6 & 85.8 & 31.4 & 30.5 & 58.5 & 24.4 & 55.1 & 58.8 & 0.0 & 82.6 & 15.3 & 67.8 & 0.5 & 87.7 & 40.0 & 82.8 & 62.5 & 65.0 & 62.0 & 50.8
\\
        &CBST~\cite{chen2021semi} & 51.8 & 84.6 & 34.9 & 47.1 & 37.5 & 29.5 & 60.1 & 69.1 & 0.0 & 79.8 & 16.5 & 67.3 & \textbf{2.7} & 88.0 & 39.2 & 84.5 & 64.5 & 71.0 & 60.4 & 47.9
\\
        &LaserMix~\cite{kong2023lasermix} & 53.7 & 85.8 & 34.7 & 45.6 & \textbf{54.9} & 35.8 & 63.2 & 73.6 & \textbf{1.3} & 79.8 & 25.0 & 68.2 & 1.8 & 87.7 & 35.4 & 84.0 & 65.8 & 70.8 & 59.4 & 48.2
\\
        \cline{2-22}
        &Ours & \textbf{58.9} & 
\textbf{89.9} & 
\textbf{41.4} & 
\textbf{60.6} & 
51.0 & 
\textbf{50.4} & 
\textbf{66.7} & 
\textbf{76.8} & 
0.0 & 
\textbf{86.7} & 
\textbf{29.4} & 
\textbf{74.7} & 
1.6 & 
\textbf{89.4} & 
\textbf{53.7} & 
\textbf{88.3} & 
\textbf{68.5} & 
\textbf{75.2} & 
\textbf{63.3} & 
\textbf{51.2}
\\
        \hline
        \hline
        \multirow{6}{*}{\rotatebox[origin=c]{90}{$20\%$}} &\textit{Sup-only}~\cite{zhu2021cylindrical} & 52.1 & 86.9 & 38.0 & 39.5 & 67.3 & 29.7 & 56.5 & 69.9 & 0.0 & 79.0 & 16.0 & 66.0 & 0.3 & 87.0 & 38.6 & 84.3 & 60.6 & 66.2 & 58.8 & 45.2
\\
        \cline{2-22}
        &DiAL~\cite{unal2023discwise} & 52.8 & 85.9 & 27.9 & 41.5 & 55.5 & 33.0 & 64.1 & 72.0 & \textbf{1.2} & 81.0 & 22.5 & 67.8 & 1.2 & 89.1 & 39.9 & 82.9 & 63.7 & 66.9 & 60.5 & 46.7
\\
        &CBST~\cite{zou2018unsupervised} & 53.3 & 86.6 & 36.8 & 40.9 & 72.9 & 28.3 & 58.0 & 69.5 & 0.0 & 81.1 & 18.3 & 68.2 & 0.7 & 88.7 & 44.3 & 83.6 & 63.3 & 64.4 & 60.3 & 47.5
\\
        &CBST~\cite{chen2021semi} & 53.9 & 85.4 & 37.2 & 44.7 & 58.9 & 32.9 & 63.5 & 71.0 & 0.0 & 81.6 & 23.1 & 69.2 & 1.9 & 88.4 & 38.2 & 83.8 & 65.7 & 69.2 & 60.2 & 48.9
\\
        &LaserMix~\cite{kong2023lasermix} & 55.1 & 88.0 & 38.8 & 51.3 & 54.8 & 36.6 & 60.2 & 73.9 & 0.0 & 78.8 & 22.7 & 71.9 & 1.5 & \textbf{90.3} & 43.3 & 85.3 & 66.5 & \textbf{70.9} & 60.3 & \textbf{51.6}
\\
        \cline{2-22}
        &Ours & \textbf{60.1} & \textbf{90.7} & 
\textbf{45.1} & 
\textbf{56.7} & 
\textbf{76.1} & 
\textbf{44.2} & 
\textbf{71.1} & 
\textbf{80.9} & 
0.0 & 
\textbf{88.7 }& 
\textbf{30.5} & 
\textbf{76.6} & 
\textbf{3.8} & 
89.5 & 
\textbf{53.7} & 
\textbf{84.6} & 
\textbf{69.1} & 
65.7 & 
\textbf{65.0} & 
\textbf{51.6} \\
        \hline
        \hline
        \multirow{6}{*}{\rotatebox[origin=c]{90}{$50\%$}} &\textit{Sup-only}~\cite{zhu2021cylindrical} & 53.8 & 87.5 & 37.2 & 41.3 & \textbf{71.4} & 29.6 & 58.8 & 80.4 & 0.0 & 81.1 & 16.7 & 67.5 & 0.4 & 88.4 & 39.4 & 83.1 & 64.4 & 65.5 & 61.8 & 47.5  
\\
        \cline{2-22}
        &DiAL~\cite{unal2023discwise} & 53.9 & 86.9 & 33.6 & 46.2 & 48.9 & 33.2 & 62.8 & 77.7 & 0.0 & 82.7 & 22.8 & 68.6 & 3.2 & 89.2 & 38.6 & 83.8 & 66.4 & 68.0 & 62.3 & 48.5 
\\
        &CBST~\cite{zou2018unsupervised} & 54.5 & 87.6 & 39.5 & 36.7 & 65.9 & 35.7 & 62.8 & 78.1 & 0.0 & 82.4 & 20.4 & 69.6 & 0.1 & 88.8 & 42.3 & 84.2 & 64.0 & 67.4 & 60.1 & 50.1  
\\
        &CBST~\cite{chen2021semi} & 54.8 & 85.1 & 35.2 & 45.2 & 68.6 & 32.0 & 65.7 & 77.9 & 0.2 & 81.2 & 21.7 & 69.0 & 1.6 & 89.2 & 40.2 & 84.5 & 65.1 & 70.1 & 60.9 & 48.5
 \\
        &LaserMix~\cite{kong2023lasermix} & 56.8 & 88.0 & 40.8 & 51.6 & 63.1 & 38.4 & 61.7 & 79.9 & \textbf{2.0} & 83.1 & 26.1 & 71.2 & 2.8 & \textbf{90.1} & 41.7 & 85.9 & \textbf{69.5} & 70.5 & 63.0 & 51.6 
\\
        \cline{2-22}
        &Ours & \textbf{61.2} & \textbf{95.1} & \textbf{45.2} & \textbf{57.1} & 62.7 & \textbf{45.5} & \textbf{71.2 }& \textbf{82.1} & 0.0 & \textbf{93.1} & \textbf{42.3 }& \textbf{78.4} & \textbf{4.3} & 89.6 & \textbf{52.4} & \textbf{88.1} & 63.3 & \textbf{74.3} & \textbf{64.7} & \textbf{53.7} \\
        \hline
    \end{tabular}
    }
\end{table*}

\hfill

\clearpage
\clearpage

\end{document}